\documentclass[final,3p,times,twocolumn]{elsarticle}

\usepackage{graphicx}
\usepackage{xcolor}
\usepackage[caption=false,justification=justified]{subfig}  

\usepackage{amssymb}
\PassOptionsToPackage{hyphens}{url}\usepackage{hyperref}
\usepackage{listings}
\usepackage{amsmath}

\usepackage{xcolor}
\usepackage[normalem]{ulem}

\usepackage{hyperref}

\definecolor{codegreen}{rgb}{0,0.6,0}
\definecolor{codegray}{rgb}{0.5,0.5,0.5}
\definecolor{codepurple}{rgb}{0.58,0,0.82}
\definecolor{backcolour}{rgb}{0.95,0.95,0.92}
\definecolor{darkGreen}{RGB}{0,110,0}

\lstdefinestyle{mystyle}{
  backgroundcolor=\color{backcolour}, commentstyle=\color{codegreen},
  keywordstyle=\color{magenta},
  numberstyle=\tiny\color{codegray},
  stringstyle=\color{codepurple},
  basicstyle=\ttfamily\footnotesize,
  breakatwhitespace=false,         
  breaklines=true,                 
  captionpos=b,                    
  keepspaces=true,                 
  numbers=none,                    
  numbersep=5pt,                  
  showspaces=false,                
  showstringspaces=false,
  showtabs=false,                  
  tabsize=2,
  framexleftmargin=-5mm,
  framexrightmargin=+0mm,
  framextopmargin=1mm,
  framexbottommargin=-3mm,
  xleftmargin=-.4cm,
  xrightmargin=+0.08cm,
}

\usepackage{float}

\newcounter{bla}

\journal{Patterns, published article available \href{https://doi.org/10.1016/j.patter.2022.100589}{here}}

\newcommand{\dfont}[1]{{\fontfamily{lmss}\selectfont
#1}}

\usepackage{balance}

\begin{document}

\begin{frontmatter}



\title{\texorpdfstring{\vspace{-2cm}}{} \dfont{DADApy}: Distance-based Analysis of DAta-manifolds in Python}


\author[a,b]{Aldo Glielmo\corref{author}}
\author[a]{Iuri Macocco}
\author[a]{Diego Doimo}
\author[a]{Matteo Carli}
\author[a]{Claudio Zeni}
\author[a]{\\Romina Wild}
\author[c,d]{Maria d'Errico}
\author[e]{Alex Rodriguez}
\author[a,e]{Alessandro Laio\corref{author}}

\cortext[author] {Corresponding authors.\\\textit{E-mail addresses:} aldo.glielmo@bancaditalia.it, laio@sissa.it}
\address[a]{International School for Advanced Studies (SISSA), Via Bonomea 265, Trieste, Italy}
\address[b]{Banca d’Italia, Italy$^{\text{**}}$\corref{disclaimer}}
\address[c]{Functional Genomics Center, ETH Zurich / UZH, Winterthurerstrasse 190, Zurich, Switzerland}
\address[d]{Swiss Institute of Bioinformatics, Quartier Sorge – Batiment Amphipole 1015 Lausanne, Switzerland}
\address[e]{The Abdus Salam International Centre for Theoretical Physics (ICTP), Strada Costiera 11, Trieste, Italy}

\cortext[disclaimer] {The views and opinions expressed in this paper are those of the authors and do not necessarily reflect the official policy or position of Banca d’Italia.}

\begin{abstract}

\dfont{DADApy} is a python software package for analysing and characterising high-dimensional data manifolds.
It provides methods  for estimating the  intrinsic dimension and the probability density, for performing density-based clustering, and for comparing different distance metrics. 
%
%
We review the main functionalities of the package and exemplify its usage in a synthetic dataset and in a real-world application.
\dfont{DADApy} is freely available under the open-source Apache 2.0 license.

\end{abstract}

\end{frontmatter}

\newpage

\section{Introduction}
\label{sec:intro}

\begin{figure*}
\includegraphics[width=0.99\textwidth]{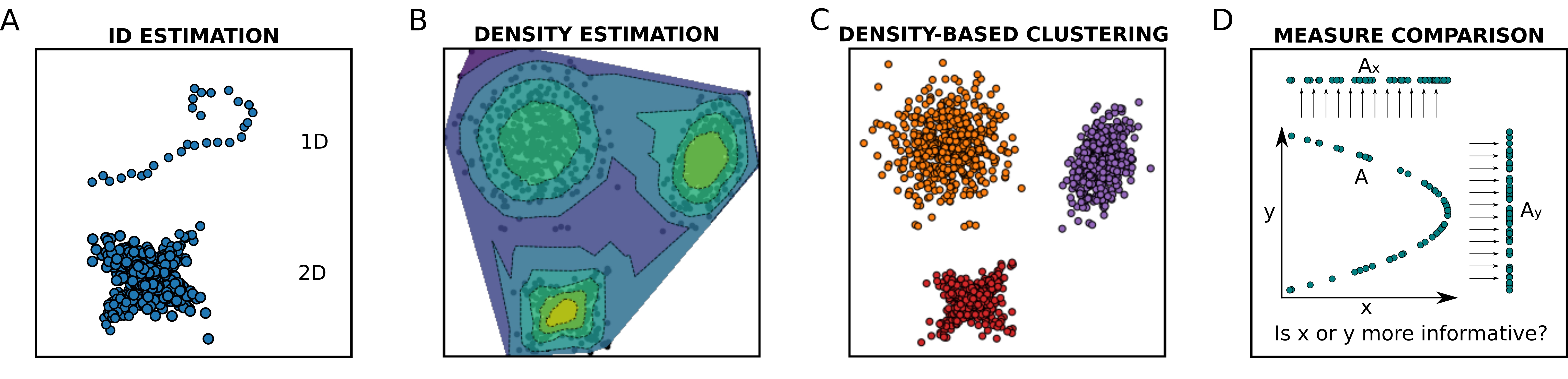}
	\caption{\textbf{An illustration of the four main classes of tasks that \dfont{DADApy} can perform}. From \textbf{A} to \textbf{D}: Intrinsic dimension estimation, density estimation, density peaks estimation (i.e., density-based clustering algorithms), and comparison of distance measures.
	}
	\label{fig:package_toolkit}
\end{figure*}

The necessity to analyse large volumes of data is rapidly becoming ubiquitous in all branches of computational science, from quantum chemistry, biophysics and materials science \cite{schutt2020machine_quantum,glielmo2021unsupervised} to astrophysics and particle physics \cite{carleo2019machine}.

In many practical applications, data come in the form of a large matrix of features,
%
and one can think of a dataset as a cloud of points living in the very high dimensional space defined by these features.
The number of features for each data point can easily exceed the thousands, and if such a cloud of points were to occupy the entire space uniformly, there would be no hope of extracting any kind of usable information from data~\cite{Keogh2010,aggarwal2001surprising}.
Luckily this never happens in practice, and real world datasets possess a great deal of hidden intrinsic structure.
The most important one is that the feature space, even if very high dimensional, is very sparsely populated. 
In fact, the points typically lie on a \emph{data manifold} of much lower dimension than the number of features of the dataset (Figure~\ref{fig:package_toolkit}A). 
A second important hidden structure which is almost ubiquitous in real world data is that the density of points on such a manifold is far from uniform (Figure~\ref{fig:package_toolkit}B). 
The data points are instead often grouped in density peaks (Figure~\ref{fig:package_toolkit}B-C), at times well separated from each other, at times organised hierarchically in ``mountain chains''.

\dfont{DADApy} implements in a single and user friendly software a set of state-of-the-art algorithms to characterise and analyse the intrinsic manifold of a dataset.
In particular, \dfont{DADApy} implements algorithms aimed at estimating the \emph{intrinsic dimension} of the manifold (Figure~\ref{fig:package_toolkit}A) and the \emph{probability density} of the data (Figure~\ref{fig:package_toolkit}B), at inferring the topography and the relative position of the density peaks by \emph{density-based clustering} (Figure~\ref{fig:package_toolkit}C) and, finally, at \emph{comparing different metrics}, finding this way the features which are better suited to describe the manifold (Figure~\ref{fig:package_toolkit}D).

%
All these approaches belong to the class of \emph{unsupervised methods} and are designed to work also in situations in which only the \emph{distances} between data points are available instead of their features.
Therefore, the same tools can be used for analysing a molecular dynamics trajectory (where features are available) but also a metagenomics or a linguistic database, where one can only define a similarity or a distance between the data.

Another important feature of the methods included in the package is that they are specifically designed in order to work even when the intrinsic dimension of the data manifold is relatively high, of order ten or more, and if the manifold is topologically complex, and, in particular, not isomorphic to a hyperplane.
Therefore, the package can be considered complementary to other packages, such as Scikit-learn \cite{ScikitLearn_ref}, which implement classical approaches for unsupervised manifold learning which should be preferred in simpler cases, such as PCA~\cite{abdi2010principal}, kernel-PCA~\cite{scholkopf1997kernel} or Isomap~\cite{balasubramanian2002isomap}.

In the following, we first briefly describe the four classes of algorithms implemented in \dfont{DADApy}.
We then illustrate the structure of the package and demonstrate its usage for the analysis of both a synthetic and a realistic dataset. 
We will also discuss the computational efficiency of the implementations, demonstrating that the package can be used to analyse datasets of $10^6$ points or more, even with moderate computational resources.


\section{Description of the methods}
\label{sec:description}

\subsection{Intrinsic dimension estimators}
\label{subsec:description_idestimation}

The \emph{intrinsic dimension} (ID) of a dataset can be defined as the minimum number of coordinates which are needed in order to describe the data manifold without significant information loss~\cite{Campadelli2015_Intrinsic_dimension,Camastra2016_Intrinsic_dimension}.
In our package we provide the implementation of a class of approaches which are suitable to estimate the ID using only the distances between the points, and not the features. 
Most of these approaches are rooted in the observation that in a uniform distribution of points, the ratio $\mu_i$ of the distances of two consecutive nearest neighbours of a point $i$ are distributed with a Pareto distribution which depends only on the intrinsic dimension. 
This allows defining a simple likelihood for the $N$ observations of $\mu_i$, one for each point of the dataset:
\begin{equation}
p( \{ \mu_i \} \mid \mathrm{ID}) = \prod_{i=1}^N \mathrm{ID} \, \mu_i^{-(\mathrm{ID}+1)}.
\label{eq:id_estimation}
\end{equation}

The ID is then estimated either by maximising the likelihood~\cite{Levina_MLE_ID}, by Bayesian inference~\cite{denti2021distributional}, or by linear regression after a suitable variable transformation~\cite{facco2017estimating}. 
We refer to these estimators as \emph{Two nearest neighbours} (2NN) estimators.

It is possible that the data manifold possesses different IDs, depending on the scale of variations considered. 
For example, a spiral dataset can be one-dimensional on a short scale, but two-dimensional on a larger scale.
Hence, one might be interested in computing an ID estimate as a function of the scale. 
The package provides two routines to perform this task.
The first method allows to probe the ID at increasing length scales by sub-sampling the original dataset. 
By virtue of the reduced number of points considered, the average distance between them will be larger; this can be then interpreted as the length scale at which the ID is computed.
Obviously, subsampling the dataset also increases the variance of the ID estimate.
The second method, an algorithm called \emph{Generalised ratios id estimator} (Gride), circumvents this issue by generalising the likelihood in Eq.~(\ref{eq:id_estimation}) to directly probe longer length scales without subsampling~\cite{denti2021distributional}.

After using one of these algorithms, one can select the ID of the dataset as the estimate that is most consistently found across different scales.
However this choice is often not straightforward, and for a more in depth discussion on this topic we refer to~\cite{facco2017estimating,denti2021distributional}

ID estimation has been successfully deployed in a number of applications, ranging from the analysis of deep neural networks \cite{NEURIPS2019_cfcce062}, to physical applications such as phase transition detection \cite{mendes2021unsupervised} and molecular force-field validation \cite{capelli2021data}.

\subsection{Density estimators}
\label{subsec:description_densityest}

The goal of density estimation is to reconstruct the  probability density $\rho(x)$ from which the dataset has been harvested.
The package implements a non-parametric density estimator called \emph{Point-adaptive $k$NN} (PA$k$)~\cite{rodriguez2018computing}, which uses as input only the distances between points and, importantly, is designed  to work under the explicit assumption that the data are contained in an embedding manifold of relatively small dimension.
This algorithm is an extension of the standard $k$NN estimator~\cite{knn_original_article}, which estimates the density on a point as proportional to the empirical density sampled in its immediate surrounding.
More precisely, the $k$NN estimates can be written as
\begin{equation}
    \rho_i = \frac{1}{N} \frac{k}{V_{i,k}},
\end{equation}
where $k$ is the number of nearest neighbours considered, and $V_{i,k}$ is the volume they occupy.
The volume is typically computed as $V_{i,k} = \omega_{\mathrm{ID}}d_{i,k}^{\mathrm{ID}}$, where $\omega_{\mathrm{ID}}$ is the volume of unit sphere in $\mathbb{R}^\mathrm{ID}$ and $d_{i,k}$ is the distance between point $i$ and its $k$th nearest neighbour.

In PA$k$ the number of neighbours $k$ used for estimating the density around point $i$ is chosen adaptively for each data point by an unsupervised statistical approach in such a way that the density, up to that neighbour, can be considered approximately constant.
This trick dramatically improves the performance of the estimator in complex scenarios, where the density varies significantly at short distances~\cite{rodriguez2018computing}.
Importantly, the volumes which enter the definition of the estimator are measured in the low-dimensional intrinsic manifold rather than in the full embedding space. 
This  prevents the positional information of the data from being diluted on irrelevant directions orthogonal to the data manifold. 
Assuming that the data manifold is Riemannian, namely locally flat, it can be locally approximated by its tangent hyperplane and distances between neighbours, the only distances used in the estimator, can be measured in this low-dimensional Euclidean space.
This allows to operate on the intrinsic manifold without any explicit parametrisation. 
The only prerequisite is an estimate of the local intrinsic dimension, since this is needed to measure the volumes directly on the manifold.

Another key difference between $k$NN and PA$k$ estimators is that $k$NN assumes the density to be exactly constant in the neighbourhood of each point, while PA$k$ possesses an additional free parameter that allows to describe small density variations.
The PA$k$ density estimator can be used to reconstruct free energy surfaces, especially in high dimensional spaces \cite{rodriguez2018computing,Zhang2018PhysRevLett,Marinelli2021,Salahub2022}, and it can also be used for a detailed analysis of the data like in \cite{HDfluctWater}, where a distinct analysis of the data points with different densities lead to some physical insight about the system under study.

The same estimator can be used also for estimating the density on points which do not belong to the dataset~\cite{Carli2021_statistically_unbiased}, a procedure that has been recently used to quantify the degree to which test data are well represented by a training dataset~\cite{zeni2021machine}.

Finally, PA$k$ is commonly used within the density-based clustering algorithms discussed in the following section.

\subsection{Density peak clustering}
\label{subsec:description_densityclustering}

The different ``peaks'' of the probability density can be considered a natural partition of the dataset into separate groups or ``clusters''.
This is the key idea underlying \emph{Density peak} (DP) clustering~\cite{rodriguez2014clustering}, implemented in \dfont{DADApy}.
This algorithms works by first estimating the density $\rho_i$ of all points $i$, for example using  the PA$k$ method described in the previous section.
Then, the minimum distance $\delta_i$ between point $i$ and any other point with higher density is computed as
\begin{equation}
    \delta_i = \min_{j \; \mid \; \rho_j > \, \rho_i} \!d_{ij}.
\end{equation}

\begin{figure*}
    \centering
    \includegraphics[width=0.99\textwidth]{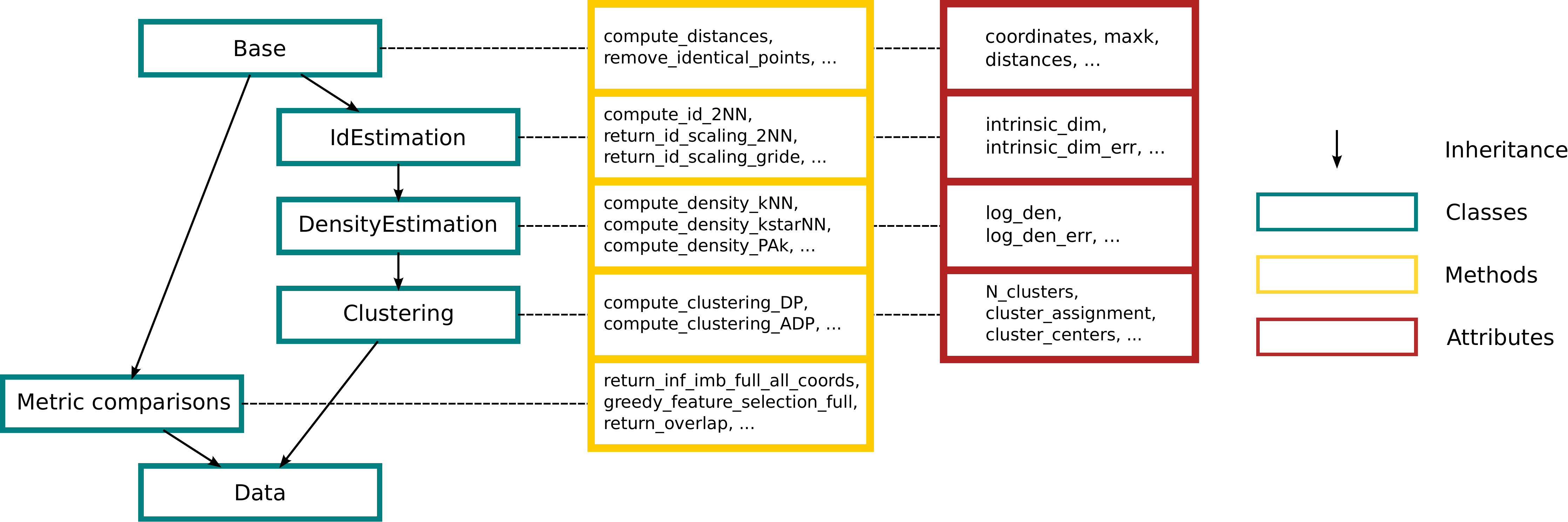}
    \caption{\textbf{The class structure of the package}. Classes are highlighted in blue boxes, and the main methods and and attributes of each class are reported in the yellow and red boxes respectively. Relationship of inheritance are indicates as black arrows. The class \textsc{Data} inherits from all other classes, thus providing easy access to all available algorithms of the package.}
    \label{fig:class_structure}
\end{figure*}

The peaks of the density (and hence the cluster centres) are expected to have both a high density $\rho_i$ and a large distance $\delta_i$ from points with higher density, and are hence selected as the few points for which both $\rho_i$ and $\delta_i$ are very large.
The selection is typically done by plotting $\rho_i$ against $\delta_i$ and visually identifying the outliers of the distribution.
Once the cluster centres are found, each remaining point is assigned to the same cluster as its nearest neighbour of higher density.

In DP clustering the density peaks must be specified by the user, and this arbitrariness represents an obvious source of errors.
The \emph{Advanced density peaks} (ADP) clustering approach~\cite{derrico2021automatic}, also available in \dfont{DADApy}, proposes a solution to this problem.
In ADP clustering, all local maxima of the density are initially considered density peaks, and a statistical significance analysis of each peak is subsequently performed.
A peak $c$ is considered statistically significant only if the difference between the log density of the peak $\ln \rho_c$ and the log density of any neighboring saddle point $\ln \rho_{cc'}$ is sufficiently larger than the sum of the errors on the two estimated quantities
\begin{equation}
    \ln \rho_{c} - \ln \rho_{cc'} > Z (\sigma_c + \sigma_{cc'}).
    \label{eq:peaks_statistical_significance}
\end{equation}
If this is not the case, the two peaks $c$ and $c'$ are merged into a single peak.
This process is iterated until no peak that is not statistically significant is remaining.
The parameter $Z$ appearing in Eq.~(\ref{eq:peaks_statistical_significance}) can be interpreted as the statistical significance threshold of the found peaks.
A higher value of $Z$ will give rise to a smaller number of peaks with a higher statistical significance.
Typical values range from~1~to~5.
ADP and DP are general clustering tools, and as such have been used in different fields, including single-cell transcriptomics \cite{ziegler2020sars,doi:10.1126/science.aad7038}, spike-sorting \cite{10.7554/eLife.34518,Sperry2020HighdensityNR}, word embedding \cite{wang2016semantic}, climate modelling \cite{margazoglou2021dynamical}, Markov state modelling \cite{pinamonti2019mechanism}, and the analysis of molecular dynamics simulations \cite{KwangHyok2018Water,Carli2020CandidateSimulations}, just to mention some of them.

Another clustering algorithm available in \dfont{DADApy} is \emph{k-peaks} clustering~\cite{Sormani2020}.
In short, this method is a variant of $ADP$ that takes advantage of the observation that the optimal $k_i$ is high in two cases: 1) In high-density regions due to the high concentration of points, and 2) in vast regions where the density is everywhere constant.
Therefore, the peaks in $k_i$ correspond either to peaks in density or to the centre of large regions with nearly constant density (e.g., metastable states stabilised by entropy). 
An example application of \emph{k-peaks clustering} can be found in \cite{Sormani2020}, where it was used to describe the free-energy landscape of the folding/unfolding process of a protein.

\subsection{Metric  comparisons}
\label{subsec:description_mancomparisons}

In several applications, the similarity (or the distance) between different data points can be measured using very different metrics. 
For instance, a group of atoms or molecules in a physical system can be represented by their Cartesian coordinates, by the set of their inter-particle distances, or by a set of dihedral angles, and one can measure the distance between two configuration with any arbitrary subset of these coordinates.
Similarly, the ``distance'' between two patients can be measured taking into account their clinical history, any subset of blood exams, radiomics features, genome expression measures, or a combination of those.

It might hence be useful to evaluate the relationships between all these different manners to measure the similarity between data points.
\dfont{DADApy} implements two methods for doing this: the \emph{neighbourhood overlap} and the \emph{information imbalance}. 
Both approaches use only the distances between the data points as input, making the approaches applicable also when the features are not explicitly defined (e.g. a social network, a set of protein sequences, a dataset of sentences).  

The neighbourhood overlap is a simple measure of equivalence between two representations~\cite{doimo2020hierarchical}.
Given two representations $a$ and $b$, one can define two $k$-adjacency matrices $A^a_{ij}$ and $A^b_{ij}$ as matrices of dimension $N \times N$ which are all zero except when $j$ is one of the $k$ nearest neighbours of point $i$.
The neighbourhood overlap $\chi(a, b)$ is then defined as
\begin{equation}
    \chi(a, b) = \frac{1}{N} \sum_i \frac{1}{k} \sum_j A^a_{ij} A^b_{ij}.
\end{equation}
Note that the term $A^a_{ij} A^b_{ij}$ is equal to one only if $j$ is within the $k$ nearest neighbours of $i$ both in $a$ and in $b$, otherwise it is zero.
For this reason, the neighbourhood overlap can also be given a very intuitive interpretation: it is the average fraction of common neighbours in the two representations.
If $\chi(a, b) = 1$ the two representations can be considered effectively equivalent, while if $\chi(a, b) = 0$ they can be considered completely independent.
The parameter $k$ can be adjusted to improve the robustness of the estimate but in practice this does not significantly change the results obtained as long as $k \ll N$~\cite{doimo2020hierarchical}.

In the original article \cite{doimo2020hierarchical}, the neighbourhood overlap was proposed to compare layer representations of deep neural networks and to analyse in this their inner workings.

The information imbalance is a recently introduced quantity capable of assessing the information that a distance measure $a$ provides about a second distance measure $b$ \cite{glielmo2021ranking}.
It can be used to detect not only whether two distance measures are equivalent or not, but also whether one distance measure is more informative than the other.
The information imbalance definition is closely linked to information theory and the theory of copula variables~\cite{glielmo2021ranking}.
However, for the scope of this article it can be empirically defined as
\begin{equation}
\begin{aligned}
\Delta(a \rightarrow b) &= \frac{2}{N} \langle r^b \mid r^a = 1 \rangle\\
& = \frac{2}{N^2}\sum_{i,j:\;r^a_{ij}=1}r^b_{ij}
\end{aligned}
\end{equation}
where $r^a_{ij}$ is the rank matrix of the distance $a$ between the points  (namely $r^a_{ij}=1$ if $j$ is the nearest neighbour of $i$, $r^a_{ij}=2$ if $j$ is the second neighbour, and so on). 
In words, the information imbalance from $a$ to $b$ is proportional to the empirical expectation of the distance ranks in $b$ conditioned on the fact that the distance rank between the same two points in $a$ is equal to one. 
If $\Delta(a \rightarrow b) \approx 0$ then $a$ can be used to describe $b$ with no loss of information. 

When measuring the information imbalances between two representations we can have three scenarios.
If $ \Delta(a \rightarrow b) \approx  \Delta(b \rightarrow a) \approx 0 $ the two representations are equivalent, if $ \Delta(a \rightarrow b) \approx  \Delta(b \rightarrow a) \approx 1 $ the two representations are independent, and finally if $ \Delta(a \rightarrow b) \approx 0 $ and $  \Delta(b \rightarrow a) \approx 1 $ we have that $a$ is informative about $b$ but not vice versa, therefore $a$ is more informative than~$b$.
The information imbalance allows for effective dimensional reduction since a small subset of features that are the most relevant either for the full set, or for a target property, can be identified and selected~\cite{glielmo2021ranking}.
This feature selection operation is available in \dfont{DADApy} and can be performed as a pre-processing step before the tools described in the previous sections are deployed.

The information imbalance proved successful in dealing with atomistic and molecular descriptors, either to directly perform compression \cite{glielmo2021ranking} or to quantify the information loss incurred by competing compression schemes \cite{darby2021compressing}.
In the original article \cite{glielmo2021ranking}, the information imbalance was also proposed for detecting causality in time series -with illustrative results shown on Covid-19 time series- and to analyse or optimise the layer representations of deep neural networks.


\section{Software structure and usage}
\label{sec:software}

\begin{figure}[ht]
    \label{fig:example_usage}
    \begin{lstlisting}[language=Python, frame=single]
    import numpy as np
    from dadapy import Data

    # initialise the "Data" class  
    # with a set of coordinates
    X = np.load("coordinates.npy")
    data = Data(X)
    
    # compute distances 
    # up to the 100th neighbour
    data.compute_distances(maxk = 100)
    
    # compute the intrinsic dimension 
    # using the 2NN method
    ID, ID_err = data.compute_id_2NN()
    
    # compute the density
    # using the PAk method
    den, den_err = data.compute_density_PAk()
    
    # find the density peaks using
    # using the ADP method
    clusters = data.compute_clustering_ADP()\end{lstlisting}
    \caption{\textbf{A simple \dfont{DADApy} script}.}
\end{figure}

\dfont{DADApy} is written entirely in Python, with the most computationally intensive methods being sped up through Cython.
It is organised in six main classes: \textsc{Base}, \textsc{IdEstimation}, \textsc{DensityEstimation}, \textsc{Clustering}, \textsc{MetricComparison} and \textsc{Data}.
The relationships of inheritance between these classes, as well as the main methods and attributes available in each class are summarised in Figure~\ref{fig:class_structure}.
The \textsc{Base} class contains basic methods of data cleaning and manipulation which are inherited in all other classes.
Attributes containing the coordinates and/or the distances defining the dataset are contained here.
Then, in a train of inheritance: \textsc{IdEstimation} inherits from \textsc{Base}; \textsc{DensityEstimation} inherits from \textsc{IdEstimation} and \textsc{Clustering} inherits from \textsc{DensityEstimation}.
Each of these classes contains as methods the algorithms described in the previous section, under the same name.
The inheritance structure of these classes is well motivated by the fact that to perform a density-based clustering one first needs to compute the density, and to perform a density estimation one first needs to know the intrinsic dimension, which can be estimated only if the distances are preliminarily computed.
The \textsc{MetricComparison} class contains the algorithms described in Section~\ref{subsec:description_mancomparisons} used to compare couples of representations using the distances between points.

The class \textsc{Data} does not implement any extra attribute or method but, importantly, it inherits all methods and attributes from the other classes.
As such, \textsc{Data} provides easy access to all available algorithms of the package and is the main class that is used in practice.

A typical usage of \dfont{DADApy} is reported in Figure~\ref{fig:example_usage}.
In this simple example a \textsc{Data} object is first initialised with the matrix containing the coordinates of the points shown in Figure~\ref{fig:package_toolkit}B-C, and later a series of methods are called sequentially to compute the distances, the intrinsic dimension, the density (Figure~\ref{fig:package_toolkit}B) and finally the density peaks (clusters) of the dataset (Figure~\ref{fig:package_toolkit}C).
In the example given, \textsc{Data} is initialised with a matrix of coordinates, and the distances between points are later computed.
Note that, however, the object could have been equivalently initialised directly with the distances between points, and all methods in the package would work equivalently.
This is particularly important for those applications for which coordinates are not available, but distances can be computed, such as DNA or protein sequences, or networks.

The main aim of the package is to provide user-friendly, fast and light routines to extract some of the most common and fundamental characteristics of a data manifold through solid statistical and numerical techniques.
\dfont{DADApy} offers high speed code with reduced memory consumption. 
These features are achieved by exploiting locality.
In particular, it is generally enough to compute the distances between each point and a small number of its neighbours (defined in \dfont{DADApy} by an attribute named \texttt{maxk}), and hence such distances can be computed and stored with close-to-linear time and memory requirements.

We believe that the Python interface of \dfont{DADApy} will encourage its rapid diffusion, as Python is by far the most used language in the computational science community nowadays. 
We are aware that Python is, however, a notoriously inefficient language for large scale computation.
In \dfont{DADApy} we circumvent this shortcoming by implementing all the heavy numerical routines using Cython extensions, which essentially generate C-compilable code that runs with very high efficiency (typically over two orders of magnitude faster in evaluation time than the pure Python implementation).
In this manner we are able to maintain the user friendliness of Python without sacrificing the computational efficiency of a fully compiled language.

\begin{figure}
    \centering
    \includegraphics[width=0.45\textwidth]{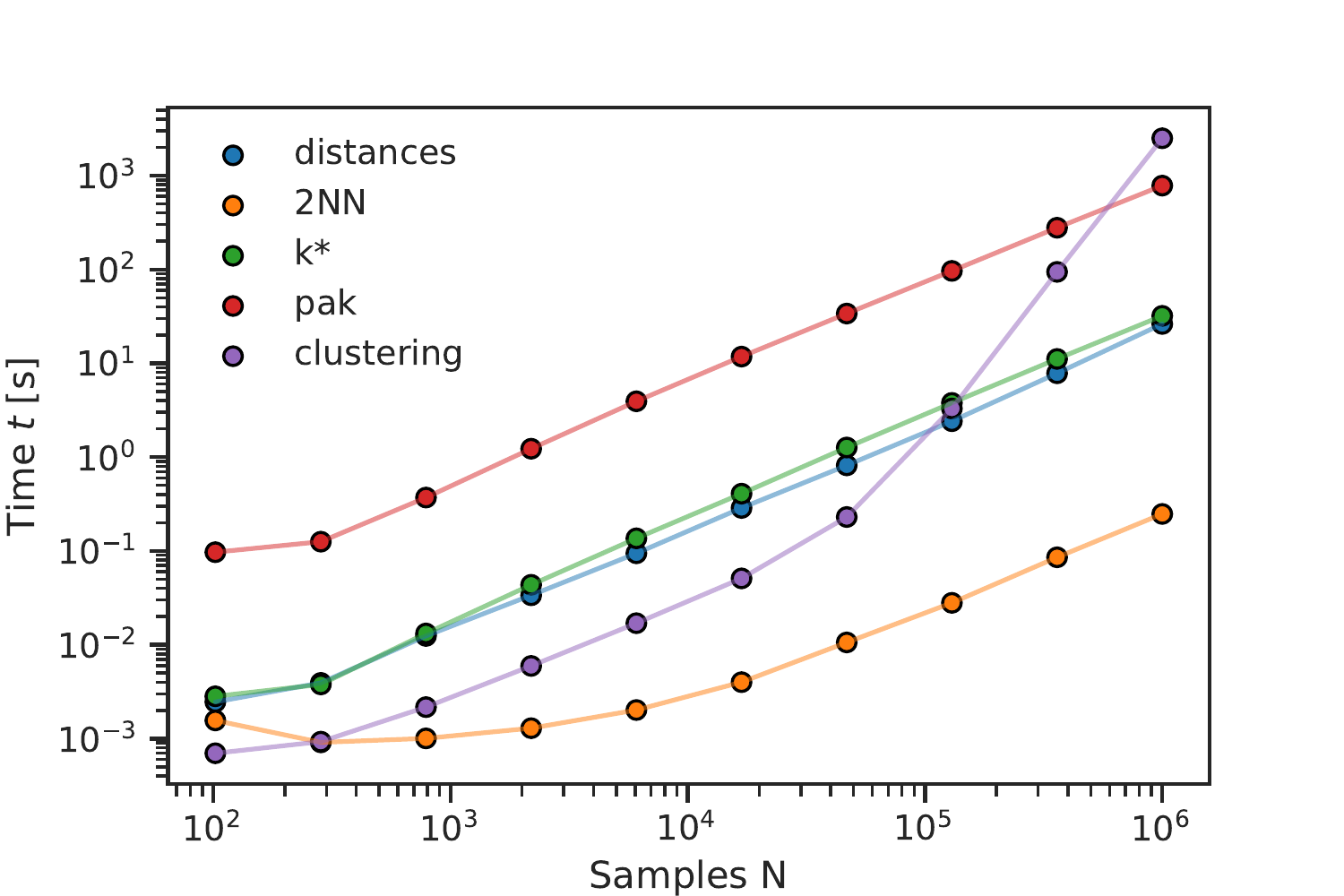}
    \caption{
    \textbf{The time complexity of \dfont{DADApy}}.
    The time required by the various routines of \dfont{DADApy} grows linearly with the number of samples $N$, with the only exception of ADP (see text for details).
    The dataset used was 2 dimensional and we set \texttt{maxk}=100. 
    The benchmark was performed on an ordinary desktop using a single Intel Xeon(R) CPU E5-2650 v2 @ 2.60GHz.
    %
    }
    \label{fig:scaling}
\end{figure}

All of the mentioned properties allow to easily analyse up to a million points on an ordinary laptop within minutes. 
This can be seen in Figure~\ref{fig:scaling}, where we report the time spent by the code on many \dfont{DADApy} routines as a function of the number $N$ of points of the dataset, using a neighbourhood of $\texttt{maxk}=100$ points.
The plot shows that all methods scale linearly in computational time with $N$, with the exception of the ADP clustering, whose scaling becomes unfavourable for more than $50$ thousand points.
This is a consequence of the neighbourhood size $\texttt{maxk}$ being much smaller than the number of points $N$ of the dataset, a condition which forces the estimation of many fictitious density peaks that take a long time to be merged together.
The problem can be solved by appropriately increasing $\texttt{maxk}$ when necessary.

The runtime performance for the computation of the distances also scales linearly with the embedding dimension $D$, while the other routines take as input the computed distances, and are thus independent on $D$.
Therefore, when $D$ is very large, say $ D \gtrsim 10^4$, the distance computation can represent the actual computational bottleneck of the package.

The code has been thoroughly commented and documented through a set of easy-to-run Jupyter notebooks, an online manual, and an extensive code reference.
This can allow new users approaching \dfont{DADApy} to quickly learn to use it, as well as to modify or extend it.

\begin{figure*}[t]
\centering
\includegraphics[width=0.95\textwidth]{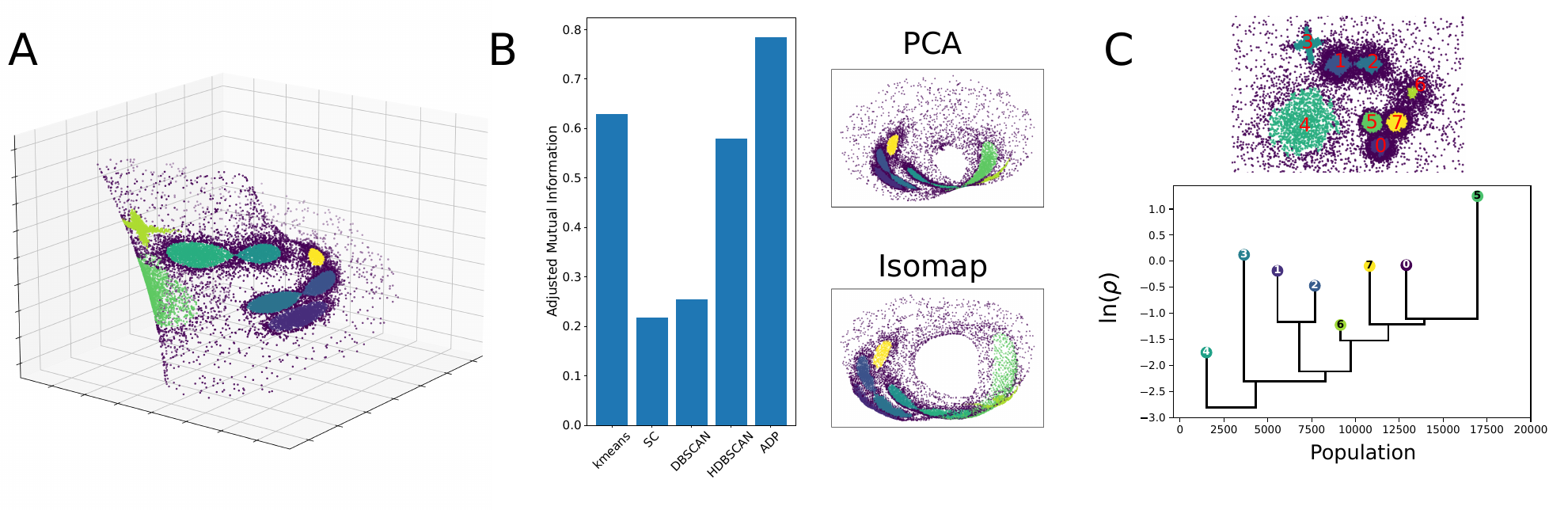}
	\caption{
	\textbf{Example usage of \dfont{DADApy} for the analysis of a topologically complex synthetic dataset}.
	Panel \textbf{A} illustrates the dataset analysed, consisting of clusters lying on a 2D sheet twisted to form a Möbius strip and immersed in a noisy 50D space. 
	Panel \textbf{B} shows the accuracy of some common clustering methods on reconstructing the original clusters, as well as two low dimensional projections.
	Panel \textbf{C}~summarises the results obtained using 2NN ID estimation, PA$k$ density estimation and ADP clustering.
	The top part shows the estimated density peaks, while the bottom part shows the dendrogram of the dataset.
    The $y$-axis of the dendrogram reports the log density of the density peaks and of the saddle points. 
    The $x$-axis provides an indication on the relative cluster sizes, since each cluster is in the middle of a region proportional to its population. 
    This region is delimited by the links in which these clusters are involved and, in the case of the first and last clusters, by the beginning and end of the graph.
	}
	\label{fig:example_results_mobius}
\end{figure*}

\begin{figure*}[t]
\centering
\includegraphics[width=0.95\textwidth]{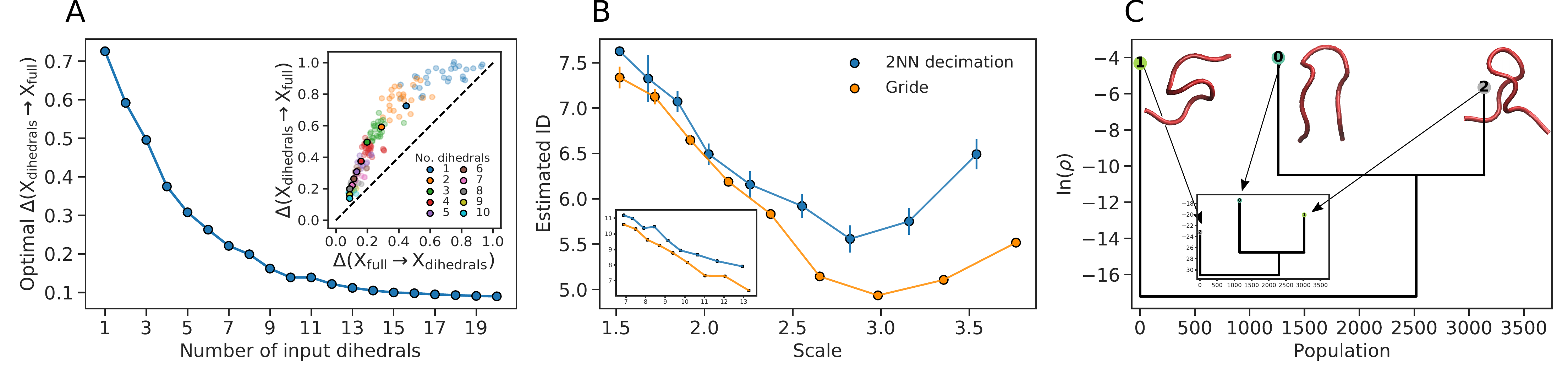}
	\caption{\textbf{Example usage of \dfont{DADApy} for the analysis of a biomolecular trajectory}.
	Panel \textbf{A} shows the computation of the information imbalance between a compact molecular representation $X_{\text{dihedrals}}$ (optimally selected sets of dihedral angles with increasing size) and a much higher dimensional one $X_{\text{full}}$ (the full space of heavy atom distances). 
	The inset shows the information imbalance between the space of heavy atom distances and the space of dihedral angles, and vice versa. For clarity, the depicted points are sparsed out. 
	Panel \textbf{B} shows the computation of the intrinsic dimension across different scales using both 2NN and Gride.
	The main graph refers to the space of 15 dihedrals, while the inset refers to the space of 4278 heavy atom distances.
	%
	Panel \textbf{C} shows a dendrogram visualisation of the peaks and the saddle points of the density, estimated using PA$k$ and the ADP clustering algorithm. 
	Peptide backbones of cluster centre structures are drawn next to their corresponding peaks.
	The main graph refers to the space of dihedrals, while the inset refers to the space of heavy atom distances.
	In both cases, the central and rightmost peaks capture the main macro states of the peptide and are much more populated than the leftmost peak.
	The two cluster assignments are identical for roughly 90\% of the data points.}
	\label{fig:example_results}
\end{figure*}

\section{Illustration on a topologically complex synthetic dataset}

We now illustrate the use of some key \dfont{DADApy} methods on the synthetic dataset depicted in Figure~\ref{fig:example_results_mobius}A, and consisting of a 2D plane with 8 clusters, twisted to form a 3D Möbius strip and finally embedded in a noisy 50D space.
The reference 2D dataset is taken from~\cite{derrico2021automatic}, and consists of data points sampled from an analytic density function, with points belonging to a single mode of this density assigned to the same cluster, and all other considered unassigned.

In spite of the 2D inner structure of the dataset, common projection methods can easily fail as a consequence of the nontrivial topological properties of the data-manifold. 
This is illustrated in Figure~\ref{fig:example_results_mobius}B, where PCA and ISOMAP projections are reported.

One key advantage of the methods implemented in \dfont{DADApy} is their ability to exploit the low dimensional structure of the data without any explicit projection.
In this case, for example, we compute the ID using the Gride method of Section~\ref{subsec:description_idestimation}, which is correctly identified around 2.
We then use the ID to provide accurate density estimates using the PA$k$ method of Section~\ref{subsec:description_densityest}, and finally identify the clusters (or density peaks) using the ADP algorithm of Section~\ref{subsec:description_densityclustering}.
The end result is a cluster assignment that is remarkably close to the ground truth, and often superior to other state-of-the-art clustering schemes that do not exploit the low dimensional structure of the data (see Figure~\ref{fig:example_results_mobius}B).

Another unique feature of \dfont{DADApy} is the ability of compactly representing the cluster structure through a special kind of dendrogram reporting the log densities of the density peaks and of the saddle points between them.
The bottom part of Figure~\ref{fig:example_results_mobius}C depicts the dendrogram for the Möbius strip data, which can be seen to provide a remarkably accurate perspective of the relationship between the estimated density peaks shown in the panel above.

Note that the dendrogram can be generated independently of the ID of the manifold, unlike most graphical data representations which are practically limited to three dimensions, thus providing a robust way to visualise the cluster structure even for the common scenario of $\mathrm{ID}>3$ manifolds.

The Jupyter notebook used to perform the analysis described in this section can be found at \url{https://github.com/sissa-data-science/DADApy/blob/main/examples/notebook_mobius.ipynb}.


\section{Usage for a realistic application}
\label{sec:results_and_discussion}

We now exemplify and showcase the usage of \dfont{DADApy} for the analysis of a biomolecular trajectory.
The dataset is composed of 41580 frames from a replica-exchange MD simulation (400 ns, 340 K replica, dt = 2 fs) of the 10-residue peptide CLN025, which folds into a beta hairpin~\cite{honda_2004}.
Several numerical representations are possible for this trajectory.
A very high dimensional one is given by the set of all distances between the heavy atoms, which amounts to 4278 features.
Such a representation is possibly very redundant, and in fact typically more compact representations are used to describe systems of this type.
For example, a compact representation for this system can be taken as the set of all its 32 dihedral angles~\cite{bonomi2009plumed, cossio2011similarity}.
In Figure~\ref{fig:example_results}A we use \dfont{DADApy} to compute the information imbalance from the space of heavy atom distances to the dihedral angles space for an increasing number of dihedral angles, and vice versa.
Not surprisingly, the compact space of dihedral angles is seen to be almost equally informative to the very high dimensional heavy atom distance space, with information imbalance $\Delta(X_{\text{dihedrals}} \rightarrow X_{\text{full}})$ lower than 0.1 when considering around $15$ angles (Figure~\ref{fig:example_results}A).
We thus select the set of the 15 most informative dihedral angles as the  collective variables to represent this dataset, since the information imbalance reaches a plateau around this number.

We then use \dfont{DADApy} to compute the ID of the dataset along different scales through both decimation and Gride algorithm \cite{denti2021distributional} (Figure \ref{fig:example_results}B).
The two procedures provide fairly overlapping estimates for the ID, which is comprised between 5 and 8 within short range distances, and thus much lower than the original feature space.
We continue by estimating the density through the PA$k$ algorithm, for which we set the ID to 7.
This ID selection is motivated by the observation that the density is a local property computed at short scales but, importantly, selecting a lower ID consistent with Figure~\ref{fig:example_results}B (say, 5 or 6) does not significantly affect the results.
Finally, we use \dfont{DADApy} to perform clustering using the ADP algorithm.
The results are shown in Figure~\ref{fig:example_results}C.

ADP clustering (Z = 4.5) produces three clusters. 
The biggest cluster is the folded beta hairpin state of the protein, as depicted in Figure~\ref{fig:example_results}C (cluster 0). 
A cluster of roughly half the size is made of a collapsed twisted loop structure (Figure~\ref{fig:example_results}C, cluster 2). 
Since CLN025 is suspected to have two main metastable states, the folded hairpin and a denatured collapsed state~\cite{mckiernan_2017}, we suggest that the twisted loop could be the dominant topology of the denatured collapsed ensemble. 
The high occurrence of the twisted loop might be due to the simulation temperature of 340 K, which is just below the experimental melting temperature of CLN025 of 343 K \cite{honda_2008}. 
Less than one percent of the structures are in cluster 1, which is composed of denatured extended and less structured topologies.

The 32-dimensional space of dihedrals used so far in our analysis is known to be well suited to differentiate meaningful protein structures, but to showcase the possibility of using \dfont{DADApy} to work in very high dimensional spaces, we performed a similar analysis also on the 4278-dimensional space of all heavy atoms distances.
Using this alternative data description we performed ID estimation with the 2NN method, density estimation with the PA$k$ estimator, and clustering with the ADP algorithm (ID = 9; Z = 3.5). 
The resulting dendrogram is shown as an inset of Figure~\ref{fig:example_results}C.
%
%

As clear from the figure, we find a remarkably similar cluster structure, defined by the two major macrostates of the molecule, the beta pin and the twisted loop, as well as the cluster with unstructured configurations.


The equivalence in the two cluster assignments is confirmed by the fact that 89\% of the data points are assigned to the same cluster independently of the data representation.

A Jupyter notebook containing the analyses performed in this Section is available at \url{https://github.com/sissa-data-science/DADApy/blob/main/examples/notebook_beta_hairpin.ipynb} along with the necessary datasets.


\section{Conclusions}
\label{sec:conclusions}

In this work we introduce \dfont{DADApy}, a software package for quickly extracting fundamental properties of data manifolds.
\dfont{DADApy} is written entirely in Python, which makes it easy to use and to extend; and it exploits Cython extensions and algorithms for sparse computation and sparse memory handling, which make it computationally efficient and scalable to large datasets.
The package is documented by a set of easy-to-run Jupyter notebooks and by a code-reference and manual available online.

\dfont{DADApy} includes state-of-the-art algorithms for intrinsic dimension estimation, density estimation, density-based clustering and distance comparison that found numerous applications in recent years, but have not yet found widespread usability.
We believe this was, at least in part, precisely due to the lack of a fast and easy-to-use software like \dfont{DADApy}, and we hope that our work will allow a growing number of practitioners from different research domains to approach the field of manifold learning.

The algorithms included in DADApy do not rely on low dimensional projections or on any strong assumptions on the structure of the data.
This can be a great atvantage, as it makes DADApy suited to analyse  topologically complex data manifolds, but it also means that DADApy cannot be used to build low dimensional maps for data visualisation.
Other shortcomings of the software are in its level of maturity for industrial-grade standards --DADApy is still a young software-- and in the relatively small number of algorithms implemented~in~it.

We plan to improve DADApy by addressing both of these issues.
On the one hand we are working on the development of algorithms that extend many of the methods discussed here, including ID estimators for discrete spaces \cite{macocco2022intrinsic}, density estimators that exploit data correlations, and more refined feature selection schemes based on the information imbalance; and intend to implement these as new DADApy methods.
On the other hand we intend to improve code quality in a variety of directions such as by increasing unit test coverage, expanding documentation and lint checks, and adding static type checking.
Finally, we will greatly welcome open source contributions to the project.

\section*{Code availability}
DADApy can be downloaded from the Github page \url{https://github.com/sissa-data-science/DADApy}.

\section*{Acknowledgments}
AG and AL acknowledge support from the European Union’s Horizon 2020 research and innovation program (Grant No. 824143, MaX `Materials design at the eXascale' Centre of Excellence).

The views and opinions expressed in this paper are those of the authors and do not necessarily reflect the official policy or position of Banca d’Italia.
%





\balance

\bibliographystyle{elsarticle-num}

\bibliography{bib}







\end{document}